\documentclass[sigconf]{acmart}
\renewcommand\footnotetextcopyrightpermission[1]{} 
\AtBeginDocument{%
  \providecommand\BibTeX{{%
    \normalfont B\kern-0.5em{\scshape i\kern-0.25em b}\kern-0.8em\TeX}}}

\setcopyright{acmcopyright}
\copyrightyear{2021}
\acmYear{2021}
\acmDOI{10.1145/1122445.1122456}

\acmConference[Chengdu '21]{Chengdu '21: ACM Multimedia}{Oct. 20--24, 2021}{Chengdu, China}
\acmBooktitle{Chengdu '21: ACM Multimedia, Oct. 20--24, 2021, Chengdu, China}
\acmPrice{15.00}
\acmISBN{978-1-4503-XXXX-X/18/06}

\acmSubmissionID{243}

\usepackage{ulem}
\usepackage{multirow}
\usepackage{booktabs}
\usepackage{times}
\usepackage{epsfig}
\usepackage{graphicx}
\usepackage{amsmath}
\begin{document}

\title{HAT: Hierarchical Aggregation Transformers for Person Re-identification}






\author{Guowen Zhang, Pingping Zhang, Jinqing Qi, Huchuan Lu$^{*}$}
\affiliation{%
  \institution{Dalian University of Technology}
  \streetaddress{No.2 Linggong Road, Ganjingzi Distra}
  \city{Dalian}
  \state{Liaoning}
  \country{China}
  \postcode{116024}}
\email{guowenzhang@mail.dlut.edu.cn;zhpp,jinqing,lhchuan@dlut.edu.cn}



\renewcommand{\shortauthors}{Guowen and Pingping, et al.}

\begin{abstract}
  Recently, with the advance of deep Convolutional Neural Networks (CNNs), person Re-Identification (Re-ID) has witnessed great success in various applications.
  However, with limited receptive fields of CNNs, it is still challenging to extract discriminative representations in a global view for persons under non-overlapped cameras.
  Meanwhile, Transformers demonstrate strong abilities of modeling long-range dependencies for spatial and sequential data.
  In this work, we take advantages of both CNNs and Transformers, and propose a novel learning framework named Hierarchical Aggregation Transformer (HAT) for image-based person Re-ID with high performance.
  To achieve this goal, we first propose a Deeply Supervised Aggregation (DSA) to recurrently aggregate hierarchical features from CNN backbones.
  With multi-granularity supervisions, the DSA can enhance multi-scale features for person retrieval, which is very different from previous methods.
  Then, we introduce a Transformer-based Feature Calibration (TFC) to integrate low-level detail information as the global prior for high-level semantic information.
  The proposed TFC is inserted to each level of hierarchical features, resulting in great performance improvements.
  To our best knowledge, this work is the first to take advantages of both CNNs and Transformers for image-based person Re-ID.
  Comprehensive experiments on four large-scale Re-ID benchmarks demonstrate that our method shows better results than several state-of-the-art methods.
  The code is released at \textcolor{red}{https://github.com/AI-Zhpp/HAT}.

\end{abstract}




\maketitle

\section{INTRODUCTION}
\begin{figure}[t]
\centering
\includegraphics[width=0.40\textwidth]{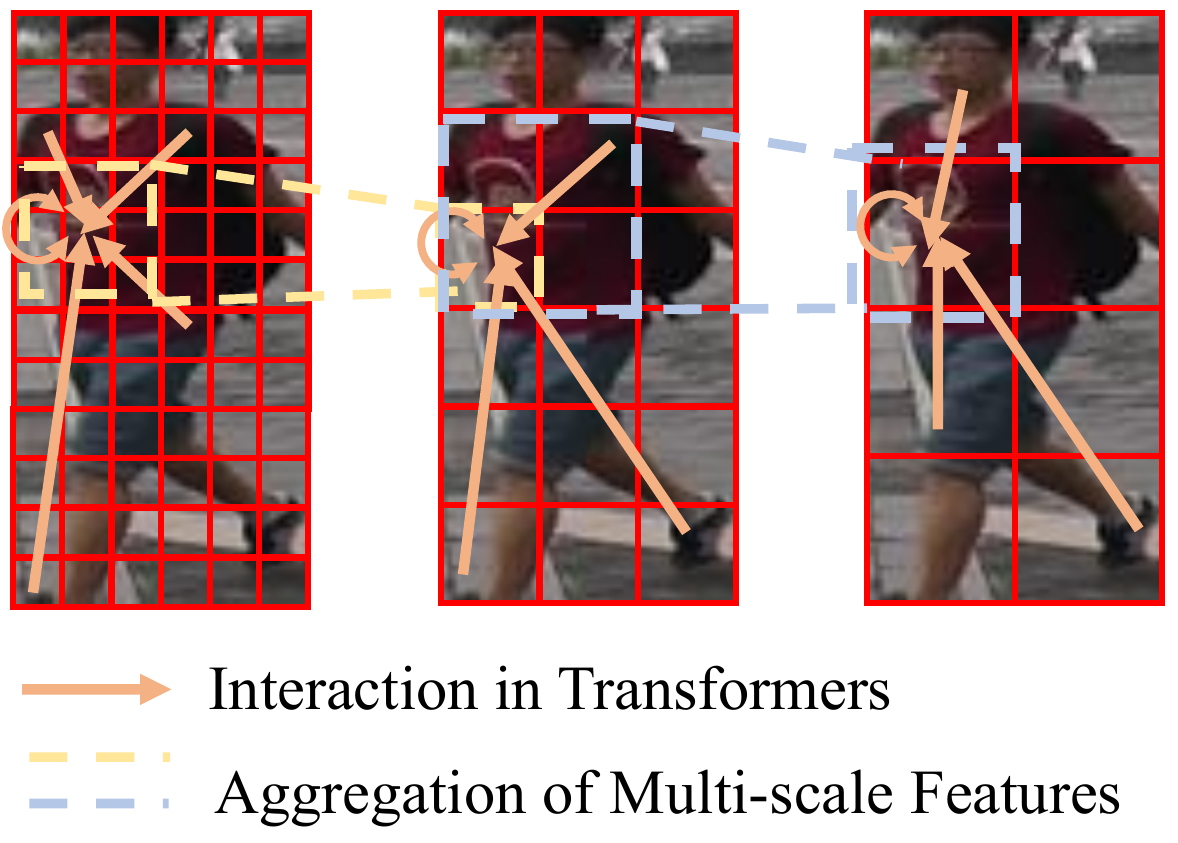}
\vspace{-4mm}
\caption{The insight of our Hierarchical Aggregation Transformer. It integrates low-level information as the global prior for enriching high-level semantic information.}
\label{figure:hat}
\vspace{-4mm}
\end{figure}
Person Re-identification (Re-ID) aims to retrieve the same person under different cameras, places and times.
As an important component of intelligent surveillance and autonomous driving, person Re-ID has drawn a surge of interests.
The challenge of person Re-ID lies in extracting rich, discriminative and robust features from person images which are under large variations, such as  occlusion, illumination, pose and background clutter.

\begin{figure*}[t]
\centering
\includegraphics[width=0.85\textwidth]{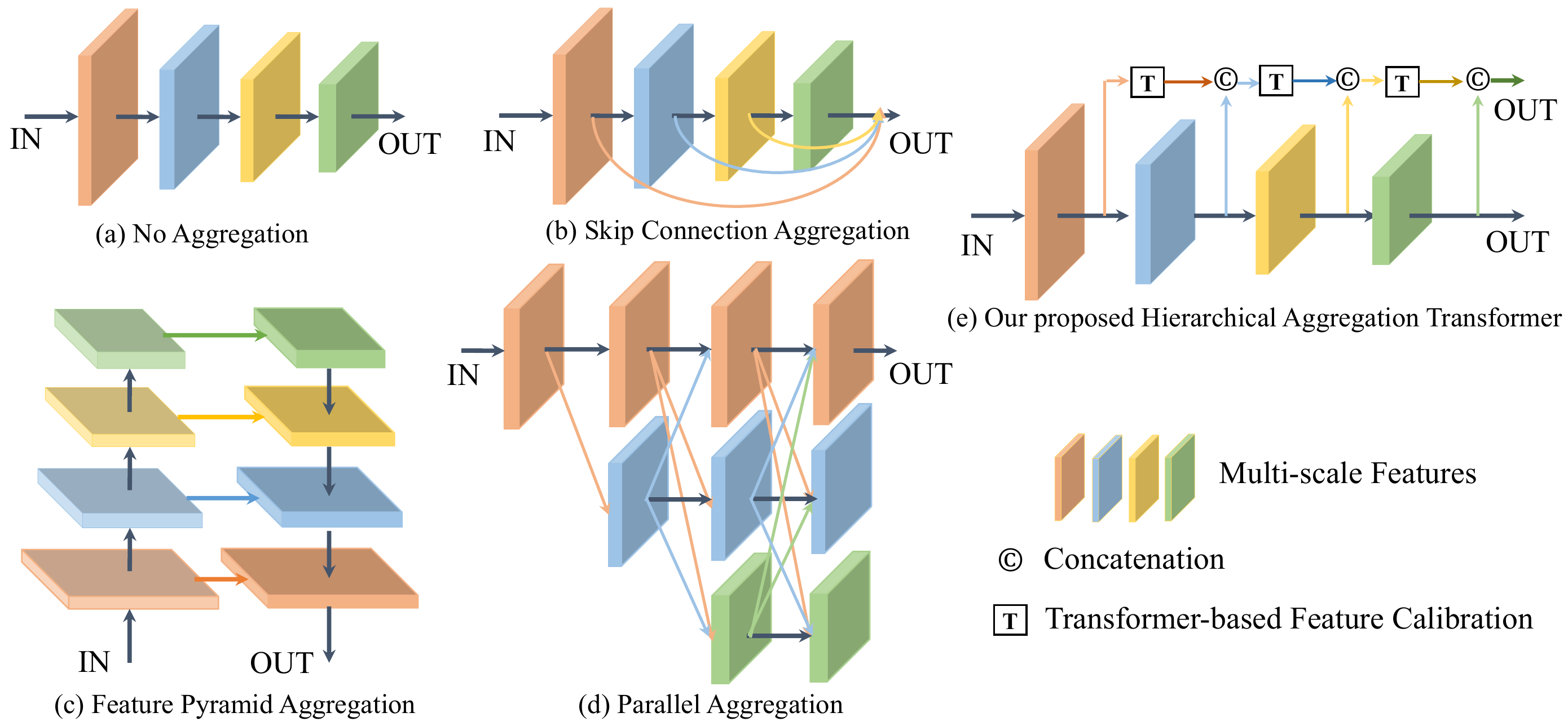}
\caption{Illustration of different aggregation approaches. (a) Backbones without aggregation. (b) Skip connection aggregation. (c) Feature pyramid aggregation. (d) Parallel aggregation. (e) Our proposed Hierarchical Aggregation Transformer, which iteratively aggregates the multi-scale features.}
\label{figure:aggregation}
\end{figure*}

Currently, with the progress of deep learning, Convolutional Neural Networks (CNNs) based person Re-ID methods have achieved great success.
The rise of CNNs as the backbone of many visual tasks such as image classification, object detection and segmentation, has encouraged researchers to explore effective structures for feature aggregation, as shown in Fig.~\ref{figure:hat}.
The trend is to design effective structures with more non-linearity, greater capacity and larger receptive field.
As shown in Fig.~\ref{figure:aggregation}(b), the utilization of skip connection aggregation is one of the most common methods to integrate the multi-level features at once.
A series of works~\cite{lin2017feature} utilize a top-down feature pyramid architecture with lateral connections to build high-level semantic feature maps as shown in Fig.~\ref{figure:aggregation} (c).

Furthermore, for image-based person Re-ID, previous works~\cite{chen2020salience,yu2017devil,liu2020hierarchical,zhou2019omni} explore the effectiveness of hierarchical features of CNNs.
For example, some works~\cite{chen2020salience,liu2020hierarchical} utilize attention-based structures to adaptively fuse multi-level features.
Zhou \emph{et al.}~\cite{zhou2019omni} propose an unified aggregation gate to dynamically fuse multi-scale features with different channel-wise weights.
Besides, some works~\cite{geirhos2018imagenet,sun2018beyond} use multi-branch structures to learn multiple granularity features from independent regions.
Those works show that fusing multi-scale features can extract more semantic information and achieve better performance.
However, previous methods may limit the performance because of the less semantic information in low-level features.
Thus, further exploration is needed on how to aggregate different level and scale features in image-based person Re-ID.

Nowadays, Transformers~\cite{vaswani2017attention} have been the de-facto standard and gains a great success in Natural Language Processing (NLP).
Recently, Transformers~\cite{he2021transreid, zhu2021aaformer,dosovitskiy2020image,carion2020end} are extend to many computer visual tasks and show superior performance.
Dosovitskiy \emph{et al.}~\cite{dosovitskiy2020image} propose the Vision Transformer (ViT) for image recognition, achieving comparable performance with traditional CNNs.
For person Re-ID, He \emph{et al.}~\cite{he2021transreid} introduce a pure transformer-based object Re-ID framework, showing state-of-the-art performance.
Zhu \emph{et al.}~\cite{zhu2021aaformer} apply an explicit alignment mechanism to enhance the capability of Transformers in person Re-ID.
The vision transformer has shown the capability of exploiting structural patterns from a global view.
Although the pure-transformer structure shows great capability and potentiality, it is limited by the need of amounts of data for pre-training.
Besides, compared with CNNs, the pure-transformer structure lacks of desirable properties such as shift, scale and distortion invariance and hierarchical structure, that make CNNs suite for visual tasks.
Inspired by CNNs and Transformers, we combine the two basic structures to maintain the merits of them for image-based person Re-ID.
Actually, CNNs are used to extract hierarchical features and the interaction in Transformers aims to aggregate the features from different scales in a global view.

In this paper, we investigate how to aggregate multi-scale features to better fuse semantic and detail information for image-based person Re-ID.
To address aforementioned issues, we propose a novel learning framework named Hierarchical Aggregation Transformer (HAT) which aggregates multi-scale features and mines the discriminative ones.
Our proposed HAT can be utilized with any hierarchical architectures and independent with the choice of backbones.
Technically, we first introduce a Deeply Supervised Aggregation (DSA) to recurrently aggregate hierarchical features from CNN backbones.
The DSA enhances and combines multi-scale features to generate more discriminative features by multi-granularity supervisions.
%
%
Inspired by the superior capability of Transformers, we propose the Transformer-based Feature Calibration (TFC) to integrate semantic information of cross-scale features.
The novel calibration module helps the model to preserve semantic and detail information from a global view.
%
%
To verify the effectiveness of our proposed methods, we conduct extensive experiments on four large-scale benchmarks.
The results clearly demonstrate the superior performance of our methods over most state-of-the-art methods.
The main contributions of our work are summarized as:
\begin{itemize}
\item We propose a novel framework (i.e., HAT) to aggregate multi-scale features to generate more discriminative features for high performance. It is the first to take advantages of both CNNs and Transformers for image-based person Re-ID.
\item We propose the Deeply Supervised Aggregation (DSA) to recurrently and adaptively aggregate the hierarchical features of backbones with multi-granularity supervisions.
\item We introduce a Transformer-based Feature Calibration (TFC) to merge multi-scale features by exploring information from a global view and promoting local information.
\item We conduct extensive ablation studies to demonstrate that our aggregation modules can effectively learn discriminative features.
Our method achieves state-of-the-art performances on four large-scale benchmark, i.e., Market1501~\cite{zheng2015scalable}, DukeMTMC~\cite{zheng2017unlabeled}, CUHK03-NP~\cite{li2014deepreid} and MSMT17~\cite{wei2018person}.
\end{itemize}
\section{RELATED WORK}
\subsection{Image-based Person Re-identification}
In recent years, with the progress of deep learning, person Re-ID has achieved great promotion in performance.
Most existing Re-ID models borrow architectures designed for image classification and other visual recognition tasks.
Generally, existing image-based Re-ID methods mainly focus on extracting discriminative features for persons.
Thus, several effective attention mechanisms are introduced to suppress irrelevant features and enhance discriminative features.
Song \emph{et al.}~\cite{song2018mask} utilize binary masks to reduce the noise of background and enhance the foreground features of persons.
Chen \emph{et al.}~\cite{chen2019mixed} capture second-order correlations of features to enhance discriminative features.
Chen \emph{et la.}~\cite{chen2019abd} integrate a pair of complementary attention modules to learn better representations.
Some works~\cite{zhang2020relation, wang2018non} capture context information from a global view to learn discriminative features.
However, it is not optimal to only focus on the global features.
Local information is also discriminative and helpful in retrieving the same person.
Su \emph{et al.}~\cite{su2017pose} utilize external priori knowledge to learn local information.
Besides, inspired by the spatial structure of human body, some works~\cite{sun2018beyond, wang2018learning} utilize horizontal partitions to learn discriminative local features.
Many recent works~\cite{chen2020salience, dai2019batch} propose to erase salient regions in deep features to mine diverse discriminative features.
Meanwhile, aggregating semantic and detail information is also an important topic for person Re-ID.
Chen \emph{et al.}~\cite{chen2020salience} use a multi-stage feature fusion block to aggregate high-level and low-level information.
Liu \emph{et al.} \cite{liu2020hierarchical} utilize correlation maps of cross-level feature-pairs to reinforce each level feature.
Zhou \emph{et al.}~\cite{zhou2019discriminative} propose the consistent attention regularizer to keep the deduced foreground mask similar from the low-level, mid-level and high-level feature maps.
Different from previous works, we extend the multi-scale aggregation operation with hierarchical and recurrent structures.
Besides, we introduce the Transformers into the aggregation operation to mine discriminative and specific information in each level.
Our aggregation method can capture long-range dependencies from a global view.
\subsection{Transformer in Vision}
Transformers~\cite{vaswani2017attention} are initially proposed in NLP and have become the new standard in many NLP tasks.
Recently, Transformers are transplanted to many vision tasks such as image classification~\cite{dosovitskiy2020image,touvron2020training}, object detection~\cite{carion2020end}, semantic segmentation~\cite{zheng2020rethinking} and visual tracking~\cite{chen2021transformer}.
Vision Transformers have shown superior capability and great potential for handling sequential data.
Carion \emph{et al.}~\cite{touvron2020training} design an end-to-end network by Transformers to remove non-maximum suppression and anchor generation.
ViT~\cite{dosovitskiy2020image} and DeiT~\cite{touvron2020training} use pure-transformers on the sequences of image patches.

For video-based person Re-ID, Liu \emph{et al.}~\cite{liu2021video} design a trigeminal network to transform video data into spatial, temporal and spatial-temporal feature spaces.
Zhang \emph{et al.}~\cite{zhang2021spatiotemporal} design perception-constrained Transformers to decrease the risk of overfitting.
For image-based person Re-ID, He \emph{et al.}~\cite{he2021transreid} utilize a pure-transformer with a side information embedding and a jigsaw patch module to learn discriminative features.
Zhu \emph{et al.}~\cite{zhu2021aaformer} add the learnable vectors of “part token” to learn part features and integrate the part alignment into the self-attention.
All those pure-transformer methods have achieved superior performance in image-based person Re-ID.
However, Transformers lack desirable properties of CNNs such as shift, scale and distortion invariance.
Besides, the hierarchical structure of CNNs can generate local spatial context at various levels.
Thus, we still keep the CNNs in our pipeline.
Inspired by superior interaction capability of Transformers (i.e., dynamic self-attention and global context), we introduce Transformers for our multi-scale feature aggregation process.
\vspace{-2mm}
\subsection{Deep Feature Aggregation}
Deep feature aggregation plays an critical role in many computer vision and multimedia tasks.
Below, we review and highlight key architectures for the aggregation of hierarchical features.
AlexNet~\cite{krizhevsky2012imagenet} emphasizes the important of deep architectures for image classification~\cite{russakovsky2015imagenet}.
To introduce deeper networks, ResNet~\cite{he2016deep} utilizes identity mappings to relieve the problem of degradation of deep networks.
GoogLeNet~\cite{szegedy2015going} shows that the auxiliary losses are helpful for network optimization.
As shown in Fig.~\ref{figure:aggregation}(c), Feature Pyramid Networks (FPN)~\cite{lin2017feature} introduces the top-down and lateral connections for equalizing resolution and propogating semantic information across different levels.
HR-Net~\cite{sun2019deep} connects the mutli-resolution subnetworks in parallel to maintain the high-resolution representations as shown in Fig.~\ref{figure:aggregation}(d).
For person Re-ID, Chen \emph{et al.}~\cite{chen2020salience} propose a skip connection aggregation to integrate aggregate low-level and high-level features as shown in Fig.~\ref{figure:aggregation}(b).
Liu \emph{et al.}~\cite{liu2020hierarchical} utilize correlation maps of cross-level features to enrich high-level features and learn salient and specific information in low-level features.
However, simple aggregation structures will lead to poor performance~\cite{yu2017devil} for Re-ID.
The challenge lies in how to keep semantic information in high-level and enrich the features with detail information in low-level at the same time.
To address these problems, we propose the hierarchical and iterative transformer-based aggregation structure.
Our method can adaptively refine and aggregate multi-level features step by step.
\section{PROPOSED METHOD}
\begin{figure*}[t]
\centering
\includegraphics[width=1.0\textwidth]{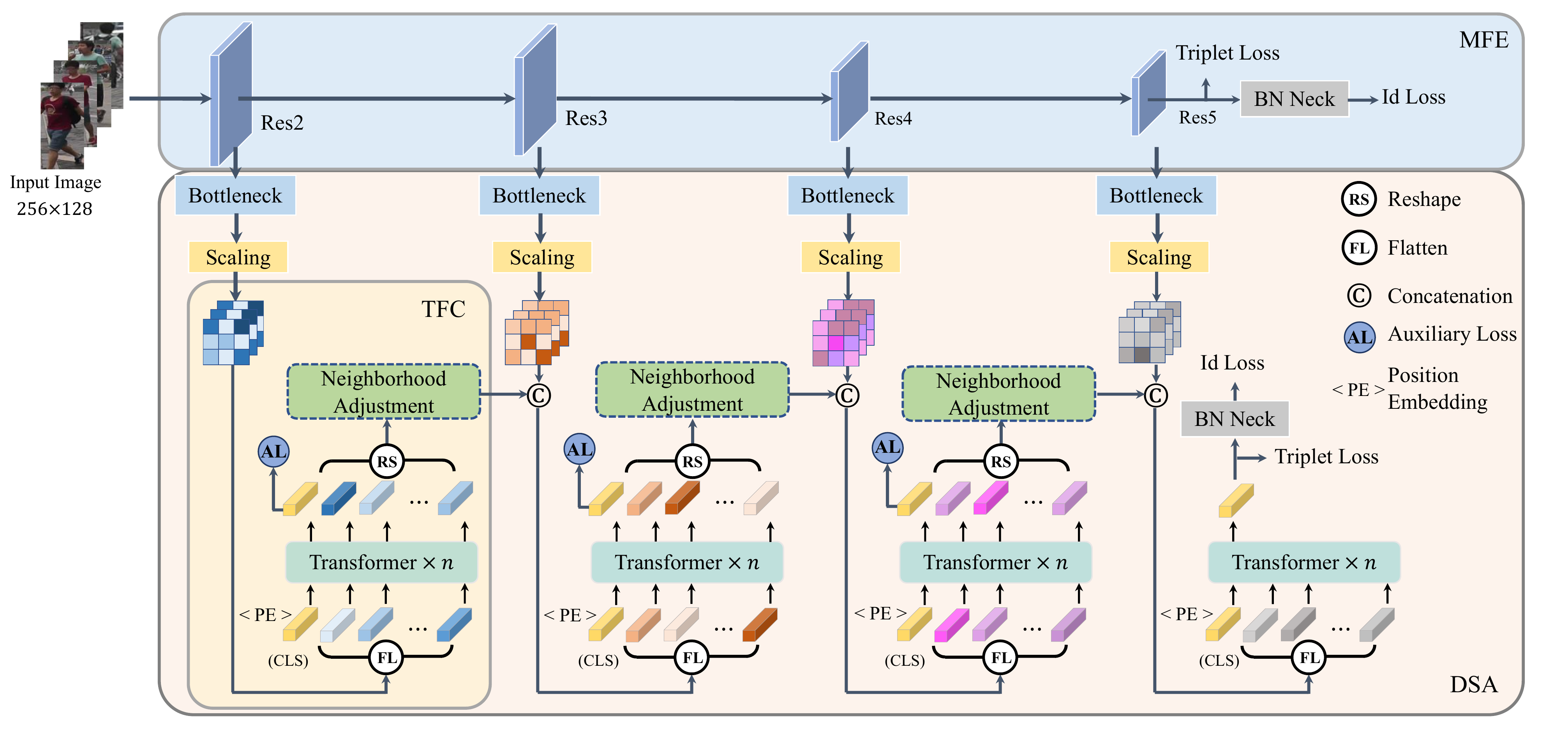}
\vspace{-6mm}
\caption{Illustration of our proposed Hierarchical Aggregation Transformer.
The person images first go forward through the ResNet-50 backbone.
After each residual block, we employ a bottleneck and scaling module to ensure the multi-level featrues have the same dimension.
Then our proposed TFC integrates low-level features as the global priors for high-level features.
Finally, multiple losses are used to supervise the whole framework in the training process.}
\label{figure:framework}
\vspace{-2mm}
\end{figure*}

In this section, we introduce the proposed Hiearchical Aggregation Transformers (HAT).
As shown in Fig.~\ref{figure:framework}, it contains three key components: a Multi-scale Feature Extractor (MFE), a Deeply Supervised Aggregation (DSA) and a Transformer-based Feature Calibration (TFC).
Our HAT is trained in an end-to-end manner.
In the following subsection, we first give an overview of the proposed HAT.
Then, we will review the basic structure of  ViT~\cite{dosovitskiy2020image}.
Finally, we will elaborate our proposed TFC and DSA in detail.
\subsection{Overview}
The proposed framework (i.e., HAT) is shown in Fig.~\ref{figure:framework}.
%
%
It can take any CNNs as backbones, such as GoogleNet~\cite{szegedy2015going} and ResNet~\cite{he2016deep}.
We adopt the widely-used ResNet-50 as our Multi-scale Feature Extractor (MFE) for its powerful feature representation for person Re-ID~\cite{chen2020salience, zhang2020relation,sun2018beyond,dai2019batch}.
To begin with, our backbone extracts hierarchical features with different scales and semantic information from Res2, Res3, Res4 and Res5.
%
%
Then, we insert the TFC module to each scale of hierarchical features.
The TFC is utilized to integrate the semantic and detail information in previous scales, and then generate global priors for next scales.
For previous- and current-level features, we utilize the concatenation operation to ensure semantic information independent of different levels before interaction.
Based on the extracted multi-scale features by TFC, the DSA utilizes multi-granularity supervisions to recurrently supervise the aggregation of the multi-scale features from low-level to high-level.
Finally, the identification loss~\cite{szegedy2016rethinking} and triplet loss~\cite{hermans2017defense} are utilized for the end-to-end training process.
In the test stage, the final outputs of backbone and HAT are concatenated for the retrieval list.
\subsection{Transformer-based Feature Calibration}
It has been demonstrated that multi-scale feature aggregation~\cite{he2016deep,yu2018deep,lin2017feature} improves the capacity of deep networks in image classification object detection and semantic segmentation .
However, person Re-ID is a special task which needs discriminative representations with sufficient semantic information.
Traditional aggregation operations of high- and low-level features will limit the performance due to less semantic information in shallower layers.
In this work, our proposed TFC aims to integrate previous level features from a global view and then aggregates previous and current level features.

Generally, given the feature maps from the s-$th$ scale block of backbones, we obtain the hierarchical feature $X_{s}\in \mathbb{R}^{C_{s}\times H\times W}$ where $C, H$ and $W$ denote the number of channel, width and height of features, respectively.
Then we employ the bottleneck which applies a stack of residual blocks~\cite{he2016deep} to transform $X_{s}$ into compact embeddings.
Most CNNs~\cite{he2016deep,xie2017aggregated} have utilized pooling layers to reduce the spatial dimension.
The pooling layers~\cite{heo2021rethinking} have been determined to be helpful to improve the model capability and generalization performance.
Besides, bilinear interpolation~\cite{GateNet} is widely used to enlarge the resolution of features.
After the bottleneck, a scaling module which is made up of a max pooling or bilinear interpolation upsampling is utilized to aggregate the resolution of hierarchical features.
The scaling module resizes the hierarchical features to the same resolution for integration and optimization.
\begin{figure}[t]
\centering
\includegraphics[width=0.30\textwidth]{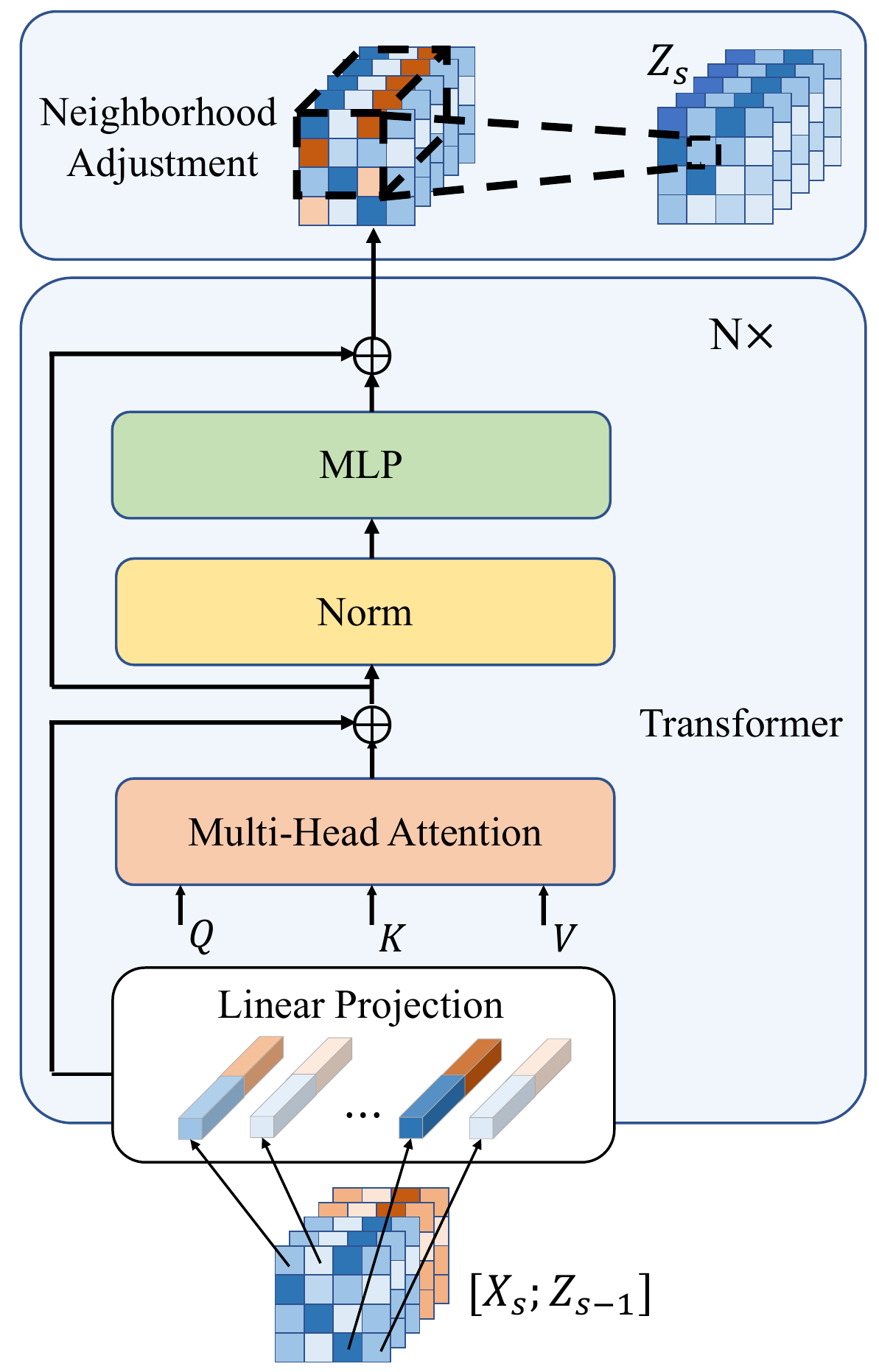}
\vspace{-4mm}
\caption{Illustration of our proposed TFC. It integrate the multi-scale features from a global view.}
\label{figure:TFC}
\vspace{-2mm}
\end{figure}

Traditional aggregation operations always utilize CNN-based structures.
However, the limited receptive fields of convolutions limit the capability of the cross-scale interaction.
Thus, for current hierarchical inputs $X_{s}$ and previous TFC output features $Z_{s-1}\in \mathbb{R}^{C_{s-1}\times H\times W}$, we use Transformers~\cite{krizhevsky2012imagenet} to strengthen and suppress information in features.
Compared with convolutions, the multi-head self-attention in Transformers can capture long-range dependencies and attend diverse information from a global view.
Meanwhile, Transformers can preserve the semantic information in interaction among different scale features.
Transformers in TFC receive a sequence of token embeddings as input.
To handle the 2D input features $Z_{s}=[X_{s};Z_{s-1}]\in \mathbb{R}^{(C_{s}+C_{s-1})\times H\times W}$,  we first flatten the feature into 2D patches $Z_{s}^{p}\in \mathbb{R}^{N\times C_{p}}$, where $N=H\times W/P^{2}$ is the number of patches, $C_{p}=C_{s}+C_{s-1}\times P^{2}$ and $P$ is set to $1$ in our framework.
Then the representation named class token (CLS) is added to the sequence, and serves as the discriminative representation.
Spatial information is incorporated by adding learnable Position Embedding (PE) for each patch, resulting in the input of sequence with a size of $Z_{s}^{p}\in \mathbb{R}^{(N+1)\times C_{p}}$.

As shown in Fig.~\ref{figure:TFC}, a Transformer is composed of a multi-head self-attention layer (MSA), a feed-forward network (FFN), layer normalizations and residual connections.
The feature $Z_{s}^{p}$ is first passed to the multi-head self-attention layer.
In each head, $Z_{s}^{p}$ is feed into three linear projections to generate query $Q\in \mathbb{R}^{(N+1)\times d}$, key $K\in \mathbb{R}^{(N+1)\times d}$,  and value $V\in \mathbb{R}^{(N+1)\times d}$, where $d=\frac{C_{p}}{N_{h}}$ and $N_{h}$ is the number of heads.
The self-attention is based on the trainable associative relation between query and key.
Then a softmax function is utilized to normalize the obtained attention weight.
The output of the multi-head self-attention is the weighted sum of $V$ by attention weights as following:
\begin{equation}
\label{eq::SA}
Attention(Q,K,V)=Softmax(\frac{QK^{T}}{\sqrt{d}})V,
\end{equation}
where $\sqrt{d}$ is utilized to normalize for numerical stability.
In this way, based on the input which is the concatenation of cross-level features, Transformers can interact the multi-level information in a global view.
With deep supervisions, Transformers can preserve the semantic information and then add the detail information mined in previous hierarchies to current level.

Then the output and input of the multi-head self-attention layer are connected by residual connections and a normalization layer,
\begin{equation}
\label{eq::MSA}
Z_{s}^{p}=LayerNorm(Z_{s}^{p}+MSA(Z_{s}^{p})).
\end{equation}
The feed-forward network (FFN) consisting of two linear projection and GELU~\cite{krizhevsky2012imagenet} activation function is applied after the MSA layer.
\begin{equation}
\label{eq::Encoder}
Z_{s}^{p}=LayerNorm(Z_{s}^{p}+FFN(Z_{s}^{p})),
\end{equation}
\begin{equation}
\label{eq::FFN}
FFN(Z_{s}^{p})=W_{2}\sigma(W_{1}Z_{s}^{p}),
\end{equation}
where $W_{1},W_{2}$ are the parameters of two linear projections and $\sigma$ is the GELU activation function.
In this way, we integrate the multi-scale features from a global view.
We replace the traditional CNN-based sturctures by Transformers to obtain a global view which is more suitable for the aggregation of multi-scale features.
And the attention-based aggregation can strengthen and suppress the multi-level information.

In Transformers, all the tokens of features are equally utilized, but the neighborhood features are more important in local regions.
To combine the advantage of CNNs (i.e., shift, scale and distortion invariance and extracting local information) with the ability of Transformers, we utilize a Neighborhood Adjustment (NeA) module after Transformers.
The output features of Transformers except CLS are reshaped to the same size as the input.
Then the features are forward into the Neighborhood Adjustment module which is composed of a stack of convolution layers with batch normalizations.
Thus, the final output of our proposed TFC is:
\begin{equation}
\begin{aligned}
\label{eq::HAToutput}
Z_{s} = Conv(Reshape(Z_{s}^{p})).
\end{aligned}
\end{equation}

\textbf{Compared with ViT~\cite{krizhevsky2012imagenet}}:
Our TFC is inspired by the recent work ViT~\cite{krizhevsky2012imagenet}.
However, it has the following fundamental differences.
First, the studied tasks are different.
ViT is designed for image classification, while our work is for image-based person Re-ID.
Second, the inputs are different.
ViT takes the tokens from previous Transformers' outputs, while our inputs are consist of hierarchical features of backbones and the outputs of previous TFC.
Third, the optimization objective is different.
ViT aims to use Transformers as backbones to learn the representation for images.
The motivation of our TFC is to merge and preserve the semantic information and detail information of multi-level features.
\begin{table*}
 \caption{Quantitative comparison on Market1501, DukeMTMC, CUHK03-NP and MSMT17 datasets.}
 \label{tab:quantable}
\centering
 \begin{tabular}{l|c|c|c|c|c|c|c|c|c|c|c|c}
 \hline
\multirow{3}{*}{Methods} & \multirow{3}{*}{Ref} &\multirow{3}{*}{Backbone}& \multicolumn{2}{c|}{\multirow{2}{*}{Market1501}}&\multicolumn{2}{c|}{\multirow{2}{*}{DukeMTMC}} &\multicolumn{4}{c|}{CUHK03-NP}&\multicolumn{2}{c}{\multirow{2}{*}{MSMT17}}\\
\cline{8-11}
~&~&~&\multicolumn{2}{c|}{~}&\multicolumn{2}{c|}{~}&\multicolumn{2}{c}{Labeled}&\multicolumn{2}{c|}{Detected}&\multicolumn{2}{c}{~}\\
\cline{4-13}
~ &~&~&mAP & Rank-1&mAP&Rank-1&mAP&Rank-1&mAP&Rank-1&mAP&Rank-1\\
\hline
\hline
DuATM~\cite{si2018dual}&CVPR18&DenseNet121&76.60&91.40&64.60&81.80&-&-&-&-&-&-\\
Mancs~\cite{wang2018mancs}&ECCV18&ResNet50&82.30&93.10&71.80&84.90&63.90&69.00&60.50&65.50&-&-\\
IANet~\cite{hou2019interaction}&CVPR19&ResNet50&83.10&94.40&73.40&83.10&-&-&-&-&46.80&75.50\\
PCB~\cite{sun2018beyond}&ECCV18&ResNet50&81.60&93.80&69.20&83.30&-&-&57.50&63.70&40.40&68.20\\
SPReID~\cite{kalayeh2018human}&CVPR18&ResNet152&83.36&93.68&73.34&85.95-&-&-&-&-&-\\
AANet~\cite{tay2019aanet}&CVPR19&ResNet152&83.41&93.93&74.29&87.65&-&-&-&-&-&-\\
CASN~\cite{zheng2019re}&CVPR19&ResNet50&82.80&94.40&73.70&87.70&68.00&73.70&64.40&71.50&-&-\\
CAMA~\cite{yang2019towards}&CVPR19&ResNet50&84.50&94.70&72.90&85.80&-&-&64.20&66.60&-&-\\
BATNet~\cite{fang2019bilinear}&ICCV19&ResNet50&84.70&95.10&77.30&87.70&76.10&78.6&73.20&76.20&56.80&79.50\\
MHN-6~\cite{chen2019mixed}&ICCV19&ResNet50&85.00&95.10&77.20&89.10&72.24&77.20&65.40&71.70&-&-\\
BFE~\cite{dai2019batch}&ICCV19&ResNet50&86.20&95.30&75.90&88.90&76.70&79.40&73.50&76.40&51.50&78.80\\
MGN~\cite{wang2018mancs}&MM18&ResNet50&86.90&\uline{95.70}&78.40&88.70&67.40&68.00&66.00&68.00&-&-\\
ABDNet~\cite{chen2019abd}&ICCV19&ResNet50&88.28&95.60&78.60&89.00&-&-&-&-&60.80&\uline{82.30}\\
Pyramid~\cite{zheng2019pyramidal}&CVPR19&ResNet101&88.20&95.70&79.00&89.00&76.90&78.90&74.80&78.90&-&-\\
JDGL~\cite{zheng2019joint}&CVPR19&ResNet50&86.00&94.80&74.80&86.60&-&-&-&-&52.30&77.20\\
OSNet~\cite{zhou2019omni}&ICCV19&OSNet&84.90&94.80&73.50&88.60&-&-&67.80&72.30&52.90&78.70\\
SNR~\cite{jin2020style}&CVPR20&ResNet50&84.70&94.40&73.00&85.90&-&-&-&-&-&-\\
RGA-SC~\cite{zhang2020relation}&CVPR20&ResNet50&88.40&\textbf{96.10}&-&-&77.40&\uline{81.10}&74.50&\textbf{79.60}&57.50&80.30\\
ISP~\cite{zhu2020identity}&ECCV20&HRNet48&\uline{88.60}&95.30&\uline{80.00}&89.60&74.10&76.50&71.40&75.20&-&-\\
CDNet~\cite{li2021combined}&CVPR21&CDNet&86.00&95.10&76.80&88.60&-&-&-&-&54.70&78.90\\
AAformer~\cite{zhu2021aaformer}&Arxiv21&ViT&87.70&95.40&\uline{80.00}&\uline{90.10}&\uline{77.80}&79.90&\uline{74.80}&77.60&\textbf{62.60}&\textbf{83.10}\\
\hline
\hline
DAT&-&ResNet50&\textbf{89.50}&95.60&\textbf{81.40}&\textbf{90.40}&\textbf{80.00}&\textbf{82.60}&\textbf{75.50}&\uline{79.10}&\uline{61.20}&\uline{82.30}\\
\hline
\end{tabular}
\vspace{-2mm}
\end{table*}
\subsection{Deeply Supervised Aggregation}

We have revisited the aggregation structures of deep networks from various perspectives.
The semantic information in high-level and detail information in low-level are both useful for visual tasks.
Many works~\cite{lin2017feature,chen2020salience, sun2019deep,GateNet} have explored effective methods to incorporate those information.
The key point is to balance the multi-level features according to the requirement of tasks.
In image-based person Re-ID, although the feature aggregation has been explored, it still has a lot of limitations.
Some works~\cite{yu2017devil, chen2020salience} have determined that simple concatenation of low-level, mid-level and high-level features will result in a worse performance for person Re-ID.
The main reason is the less semantic information of low-level features.
Based on this finding, we propose the Deeply Supervised Aggregation (DSA) for person Re-ID.
With DSA, we introduce multi-granularity supervisions to supervise the aggregation process to relieve the problem of less semantic information in low-level features.
DSA follows the iterated stacking of the backbone architecture.
We divide the stacked blocks of the MFE according to the hierarchy of backbones.
Direct concatenation of deeper and shallower layers without any correct guidance
will limit the performance of networks for person Re-ID.
Thus, we propose to progressively aggregate and refine the representation with multi-granularity supervisions by our proposed DSA.
Our DSA begins at the shallower layers with more detail and less semantic information and then recurrently merges deeper features with more semantic and less detail information.
Besides, to enhance the semantic information in the interaction of TFC, we utilize auxiliary losses to supervise the hierarchical aggregation.
The auxiliary loss is composed of the identification loss and triplet loss to keep the same optimization objective as the supervision of the whole framework.
In this way, the shallower features deliver detail information to deeper features and the semantic information in high-level can be preserved.
The DSA for hierarchical features $X_{1},X_{2},...,X_{n}$ with enhanced semantic information is formulated as:
\begin{equation}
\begin{aligned}
\label{eq::DSA}
&DSA(X_{1},...,X_{n}) =\\
&\left\{\begin{matrix}
 & X_{1} & n=1, \\
 & TFC(Concat(DSA(X_{1},...,X_{n-1}), X_{n}))&otherwise.
\end{matrix}\right.
\end{aligned}
\end{equation}
\subsection{Loss Functions}
Following previous works~\cite{luo2019bag,chen2020salience}, we treat each identity as a distinct class.
The label smoothed identification loss~\cite{szegedy2016rethinking} is used to supervise the CNN backbones and Transformers in the training procedure.
The identification loss is defined as:
\begin{equation}\label{eq::idloss}
\mathcal{L}_{id} = \sum_{i=1}^{N}-q_{i}\log(p_{i}),
\end{equation}
where $p_{i}$ is the predicted logit of identity $i$ and $q_{i}$ is the ground-truth label.
The parameter $\varepsilon$ in label smoothing is set to 0.1.
Besides, the hard triplet loss~\cite{hermans2017defense} is utilized to make that inter-class distance is less than itra-class distance.
\begin{equation}
\mathcal{L}_{tri} = [d_{pos}- d_{neg} + m]_{+},
\end{equation}
where $d_{pos}$ and $d_{neg}$ are respectively defined as the distance of postive sample pairs and negative sample pairs.
$[d]_{+}$ represents $max(0,x)$ and $m$ is the distance margin.
The auxiliary loss $\mathcal{L}_{al}$ is composed of identification loss and triplet loss with the coefficient $\lambda$ for HAT in different hierarchical levels.
Therefore, the overall loss function of the framework is:
\begin{equation}
\mathcal{L}_{reid} = \mathcal{L}_{id}+\mathcal{L}_{tri}+\lambda\sum_{i}^{n_{al}}(\mathcal{L}_{al}).
\end{equation}
where $n_{al}$ is the number of the stage blocks.
\section{EXPERIMENTS}
\subsection{Datasets and Evaluation Metrics}
To verify the effectiveness of our proposed framework, we conduct experiments on four large-scale datasets, i.e., Market1501~\cite{zheng2015scalable}, DukeMTMC~\cite{zheng2017unlabeled}, CUHK03-NP~\cite{li2014deepreid} and MSMT17~\cite{wei2018person}.
Market1501 dataset contains 19,732 gallery images and 12,936 training images captured from six cameras by Deformable Part Model (DPM) detector~\cite{felzenszwalb2008discriminatively}.
DukeMTMC-reID dataset contains 1,404 identities, 16,522 training images, 2,228 queries and 17,661 gallery images.
CUHK03 dataset contains 13,164 images of 1,467 identities and the partition of this dataset follows the method in ~\cite{zhong2017re}.
The most challenge dataset MSMT17 is the largest.
There are 126,441 images of 4,101 identities observed under 15 different camera views.
All the bounding boxes are captured by Faster R-CNN~\cite{ren2016faster}.
Following previous works, we adopt mean Average Precision (mAP) and Cumulative Matching Characteristics (CMC) at Rank-1 as our evaluation metrics.
\subsection{Implementation Details}
In our work, we uniformly resize all the input images to $256\times 128$, then followed by randomly cropping, horizontal flipping and random erasing~\cite{luo2019bag} as data augmentation.
Besides, there are $B=P\times L$ images sampled to the triplet loss and identification loss in a mini-batch for every training iteration.
We randomly select $P=16$ identities and $L=4$ for each identity.
The ResNet-50~\cite{he2016deep} pre-trained on ImageNet~\cite{karpathy2014large} is utilized as our MFE.
A warmup strategy with a linearly growing rate from $4\times 10^{-6}$ to $4\times 10^{-4}$ is used for the first 10 epoch.
The learning rate of TFC is half of the whole structure.
The learning rate begins to decrease at 50 epoch, and reduce for every 20 epoch with a factor of 0.4.
We employ Adam~\cite{kingma2014adam} as our optimizer for total 150 epoch.

\subsection{Comparison with State-of-the-arts}
\textbf{Market1501:}
Tab.~\ref{tab:quantable} shows the evaluation on the Market1501 dataset.
We compare very recent CNN-based and Transformer-based methods with our model.
From the results, we can see that our framework achieves the best performance in mAP.
Our Rank-1 is also very comparable to the best result achieved by RGA-SC which learns a global scope attention to align features.
The results indicate that our proposed framework can effectively aggregate detail information in shallower layers and semantic information in deeper layers to retrieve the hard samples.

\textbf{DukeMTMC:} As shown in Tab.~\ref{tab:quantable}, our framework achieves the best performance in mAP/Rank-1.
It is worth pointing out that the backbone of AAformer is ViT~\cite{krizhevsky2012imagenet} which is a stronger baseline than ResNet50.
%
%
The result indicates that our framework outperforms other state-of-the-art methods at least $1.0\%$ and $0.3\%$ in mAP and Rank-1 .
It indicates that our method can capture more discriminative features than other methods.

\textbf{CUHK03-NP:} Tab.~\ref{tab:quantable} shows that our HAT also achieves the best performance on this dataset.
CUHK03-NP has fewer samples, and the annotation of bounding boxes is obtained by detection~\cite{felzenszwalb2009object}.
%
%
Our framework outperforms state-of-the-art methods at least $2.2\%$, $2.7\%$ in mAP/Rank-1 for labeled CUHK03,
In detected CUHK03, our method achieves the bset mAP and the second best in Rank-1.

\textbf{MSMT17:} On this dataset, our framework achieves the second best performance in mAP and Rank-1 as shown in Tab.~\ref{tab:quantable}.
The best Rank-1 is achieved by AAformer~\cite{chen2020salience}.
The MSMT17 is a large-scale dataset which covers a long period of time and presents complex lighting variations.
AAformer uses ViT~\cite{krizhevsky2012imagenet} as the backbone which is better in capturing long-range dependencies than CNNs.
However, the performance of our framework is also comparable.
\begin{figure*}[t]
\centering
\includegraphics[width=0.94\textwidth]{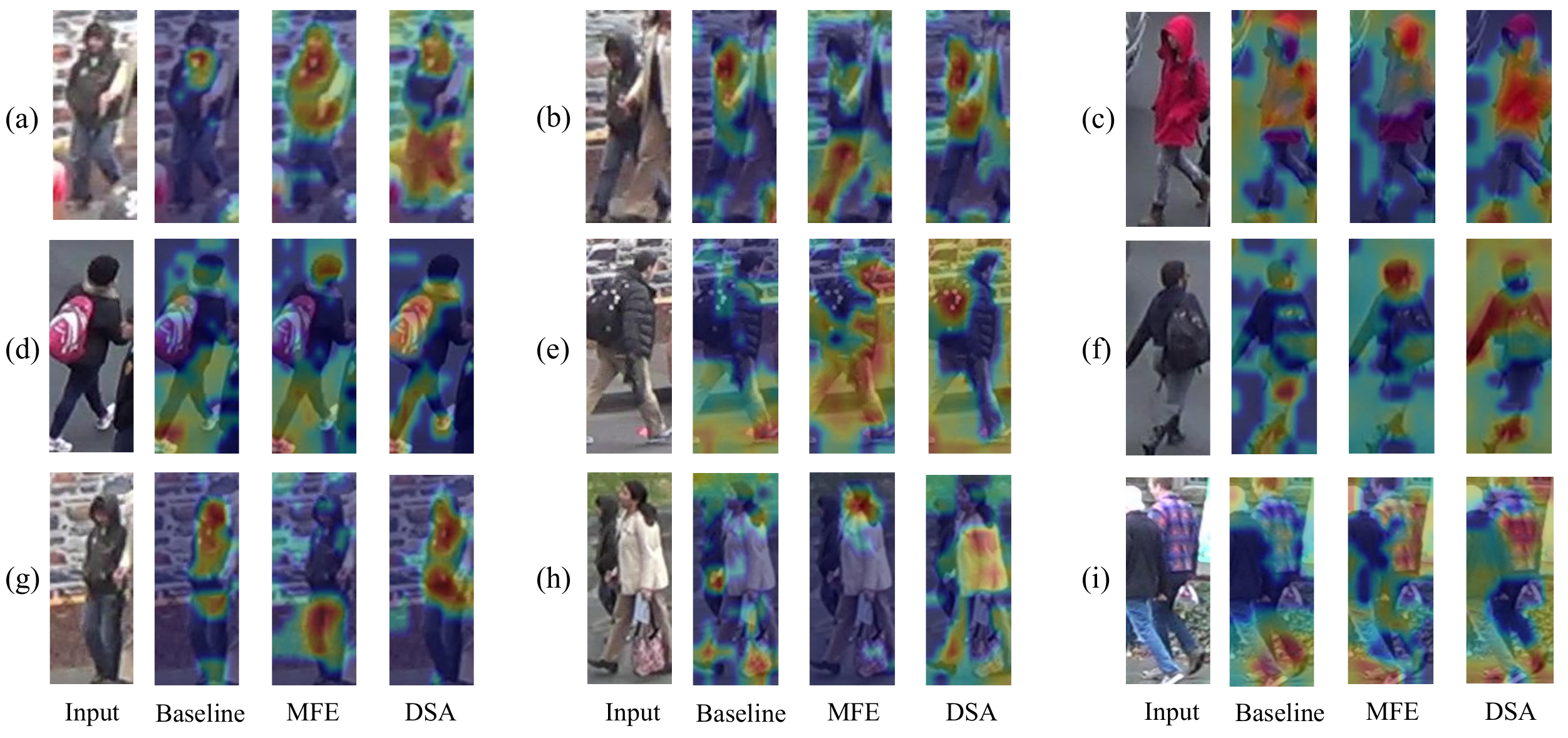}
\caption{Feature visualization of baseline and DSA and MFE of our proposed HAT.}
\label{figure:visal}
\end{figure*}
\subsection{Ablation Studies}
To demonstrate the effects of our proposed methods, we perform ablation studies on DukeMTMC dataset.
As for our CNN backbone, we use the strong baseline~\cite{luo2019bag} with ResNet50~\cite{he2016deep} as our start point.
We employ the identificaiton loss, triplet loss and auxiliary loss for the whole framework.

\textbf{TFC}: To verify the effectiveness of our proposed TFC, we compare the performance of different combinations of multi-level features.
First we respectively define the number of Transformer layers in each TFC as $\{n_{1},n_{2},...,n_{n}\}$ along the hierarchical levels.
$n_{i}=0$ means that the features at i-$th$ level are not utilized for aggregation.
To follow the setting in ViT~\cite{dosovitskiy2020image}, we fix the number of Transformers of all TFC to 12.
\begin{table}
\centering
\begin{tabular}{l|c|c|c|c}
\hline
\multirow{2}*{Method}&\multicolumn{2}{c|}{DukeMTMC}&\multicolumn{2}{c}{Market1501}\\
\cline{2-5}
\multirow{2}*{~}&mAP&Rank-1&mAP&Rank-1\\
\hline
\hline
Baseline~\cite{luo2019bag}&75.90&86.30\\
\{12,0,0,0\}&73.80&83.70&82.30&92.60\\
\{0,12,0,0\}&76.50&87.00&86.30&94.10\\
\{0,0,12,0\}&79.80&89.00&88.30&95.20\\
\{0,0,0,12\}&79.00&88.40&88.10&94.90\\
\{3,3,3,3\}&80.30&89.10&89.30&95.50\\
\{0,4,4,4\}&80.40&89.60&89.30&95.70\\
\{4,4,4,0\}&81.10&89.60&89.40&95.30\\
\{3,4,5,0\}&81.30&89.50&89.40&95.30\\
\{3,3,6,0\}&81.40&90.40&89.50&95.60\\
\hline
\{0,0,12,0\}&79.80&89.00&88.50&95.20\\
\{0,0,6,0\}&79.70&89.10&88.90&95.50\\
\hline
\end{tabular}
\newline
\caption{Ablation analysis of TFC on DukeMTMC.}
\label{tab:abaTFC}
\vspace{-6mm}
\end{table}
As shown in Tab.~\ref{tab:abaTFC}, we first conduct extensive experiments to investigate the effectiveness of the different hierarchical features.
To further explore the effectiveness of features for each aggregation level, we conduct separate experiments for each hierarchical level.
Compared with low-level and high-level features, only utilizing the mid-level features in Res4 block performs better than only utilization of others.
It indicates that the low-level features are not suitable for retrieval without sufficient semantic information.
It worth noting that the performance of utilizing the highest level features is worse than the features in Res4.
We infer that the supervisions of MFE and TFC on the highest level features will make it hard to converge and thus the networks have a worse performance.
Besides, the results also indicate that deeper features need more Transformers to integrate the multi-level information by comparing the results of $\{4,4,4,0\}$ and $\{3,3,6,0\}$.
The comparison of $\{0,0,12,0\}$ and $\{0,0,6,0\}$ determines that the accuracy tends to converge with the increase of the depth of Transformers.
The results show that our proposed aggregation methods can significantly improve the performance by $5.3\%$ and $4.1\%$ in mAP and Rank-1.
\begin{table}
\centering
\begin{tabular}{l|c|c|c|c}
\hline
\multirow{2}*{Method}&\multicolumn{2}{c|}{DukeMTMC}&\multicolumn{2}{c}{Market1501}\\
\cline{2-5}
\multirow{2}*{~}&mAP&Rank-1&mAP&Rank-1\\
\hline
\hline
Baseline~\cite{luo2019bag}&75.90&86.30\\
HAT w/o MFE Supervision&76.50&88.20&85.30&94.10\\
HAT w/o AL&80.40&89.50&88.70&95.00\\
HAT w/o NeA&81.10&90.20&89.50&95.60\\
HAT&81.40&90.40\\
\hline
\end{tabular}
\newline
\caption{Ablation analysis of HAT on DukeMTMC. AL: Auxiliary Loss. NeA: Neighborhood Adjustment.}
\label{tab:abaHATframework}
\vspace{-6mm}
\end{table}
\begin{table}
\centering
\begin{tabular}{l|c|c|c|c}
\hline
\multirow{2}*{Method}&\multicolumn{2}{c|}{DukeMTMC}&\multicolumn{2}{c}{Market1501}\\
\cline{2-5}
\multirow{2}*{~}&mAP&Rank-1&mAP&Rank-1\\
\hline
\hline
Baseline~\cite{luo2019bag}&75.90&86.30\\
d = 8&81.20&89.80&89.70&95.50\\
d = 16&81.40&90.40\\
d = 32&80.50&89.50&88.90&95.50\\
\hline
\end{tabular}
\newline
\caption{Ablation analysis of Scaling on DukeMTMC.}
\label{tab:abascale}
\vspace{-6mm}
\end{table}
\begin{table}
\centering
\begin{tabular}{l|c|c|c|c}
\hline
\multirow{2}*{Method}&\multicolumn{2}{c|}{DukeMTMC}&\multicolumn{2}{c}{Market1501}\\
\cline{2-5}
\multirow{2}*{~}&mAP&Rank-1&mAP&Rank-1\\
\hline
\hline
Baseline~\cite{luo2019bag}&75.90&86.30\\
$\lambda$ = 0&80.40&89.50&88.80&95.00\\
$\lambda$ = 0.1&80.60&89.50&89.20&95.40\\
$\lambda$ = 0.3&81.60&90.10&89.70&95.50\\
$\lambda$ = 0.5&81.40&90.40&89.80&95.80\\
$\lambda$ = 0.8&80.90&89.40&89.50&95.60\\
$\lambda$ = 1.0&80.90&89.80&89.20&95.40\\
\hline
\end{tabular}
\newline\caption{Ablation analysis of $\lambda$ of auxiliary loss on DukeMTMC.}
\label{tab:abalam}
\vspace{-8mm}
\end{table}

\textbf{HAT}: As shown in Tab.~\ref{tab:abaHATframework}, we evaluate the effectiveness of deep supervisions on DukeMTMC.
All those experiments are under the setting of $\{3,3,6,0\}$.
Without the supervision for MFE, the accuracy of our proposed HAT decreases by $4.9\%$ mAP and $2.2\%$ Rank-1.
It determines that the supervisions for backbones which are utilized for the multi-scale feature extraction is critical.
The supervision of MFE ensures the semantic information in different hierarchical features.
Besides, the auxiliary losses (AL) which are composed of multi-granularity supervisions can improve the performance of HAT by $1.0\%$ mAP and $0.9\%$ Rank-1.
The result shows the auxiliary losses can preserve the semantic information in the interaction of TFC.
Meanwhile, by adding the auxiliary losses to TFC, we can boost semantic information in the shallower layers, increase the gradient signal that gets propagated back, and provide additional regularization.
As shown in Tab.~\ref{tab:abalam}, the auxiliary loss which with a small coefficient has a better performance.
At last, the Neighborhood Adjustment (NA) module can further improve the capability of our framework by enhancing local information.
\begin{figure}[t]
\centering
\includegraphics[width=0.42\textwidth]{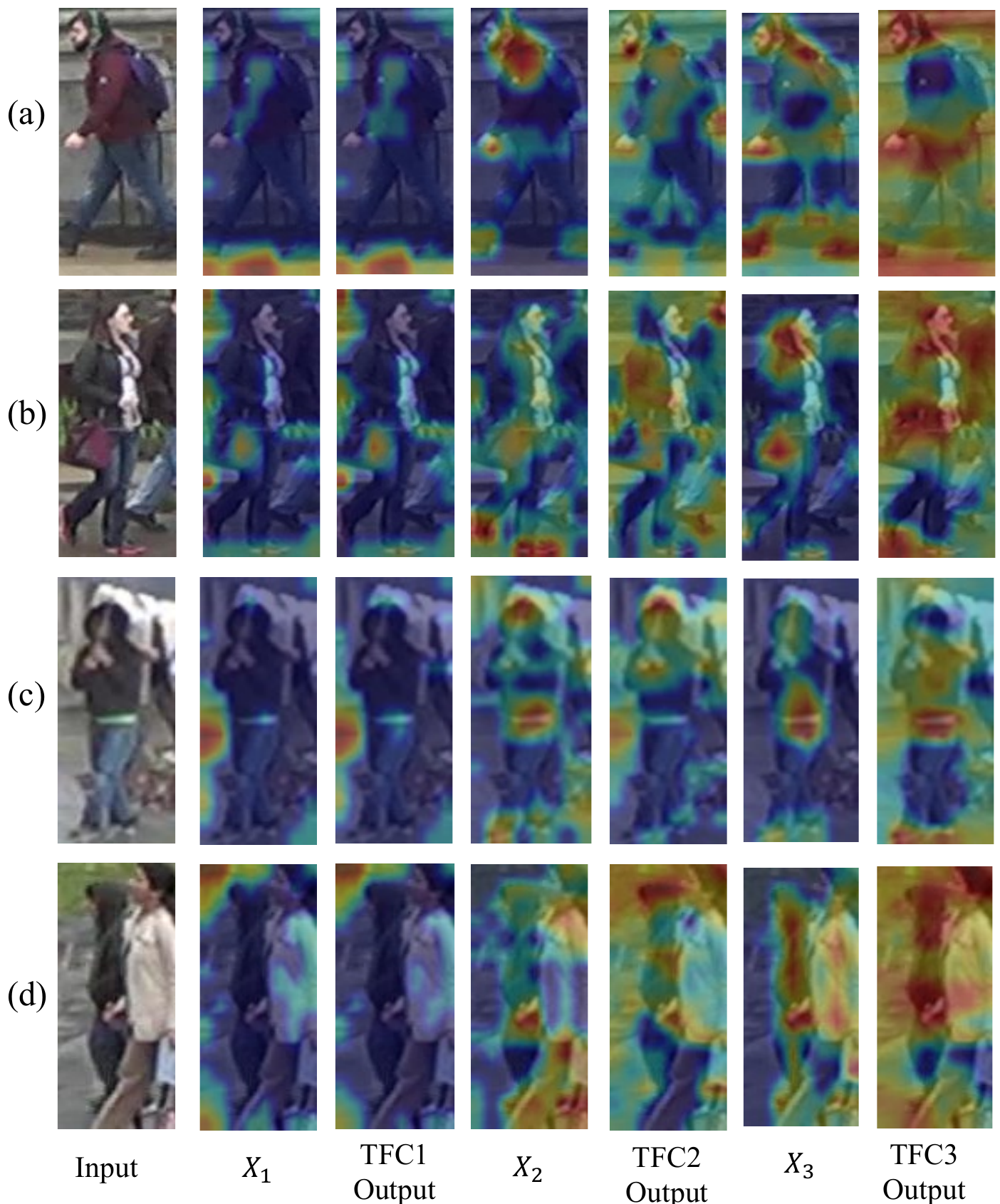}
\caption{Visualization of averaged feature maps in TFC.}
\label{figure:trans}
\vspace{-4mm}
\end{figure}

\textbf{Scaling}: To keep the resolution of hierarchical features consistent, the scaling module resizes the resolution of multi-scale features into $(\frac{H}{d},\frac{W}{d})$.
For computational and space complexity, we conduct experiments with $d=8,16$ and $32$ to verify the effectiveness of scaling.
As shown in Tab.~\ref{tab:abascale}, the resolution with $d=16$ has a better performance than $d=8$.
It shows that a larger resolution may capture more information in features for interaction in TFC.
Besides, the accuracies of $d=8$ and $d=16$ are close.
No improvement with increased resolution may be the introduction of more irrelevant features in the aggregation.
\subsection{Visualization}
As shown in Fig.~\ref{figure:visal}, we present examples of different identities and their CAM~\cite{zhou2016learning} visualization of MFE and DSA.
To verify the effectiveness of our proposed HAT, we compare the CAM visualization of the baseline with our methods.
In each example, from the left to right are the original image, the visualization of the baseline, MFE and DSA features.
It can be observed that our aggregation method can capture more detail information than the baseline.
For example, as shown in Fig.~\ref{figure:visal}(d) and (e), the feature maps of baseline only focus on the shoes of the person.
With aggregating the multi-scale features by our HAT, the networks tend to focus on the pattern of bags and clothes.
%
%
Besides, the averaged feature maps of different TFC from TFC (Res2) to TFC (Res4) are shown in Fig.~\ref{figure:trans}.
The useful information increases from low-level to high-level by aggregating multi-scale features.
The results determine that our proposed HAT can mine the discrminative features from a global view and thus capture more useful information.
From the visualization, one can find that our methods can effectively capture more detail information such as texture to improve the performance.
\section{CONCLUSION}
In this paper, we investigate how to effectively aggregate multi-scale features to better fuse semantic and detail information for image-based person Re-ID.
To this goal, we propose a novel framework named HAT consisting of TFC and DSA.
The proposed TFC can merge and preserve the semantic information and detail information in cross-level features.
It integrates low-level detail information as the global prior for high-level semantic information.
Besides, DSA can iteratively aggregate the hierarchical features of backbones with multiple losses.
Eventually, extensive experiments on four large-scale benchmarks demonstrate that our method achieves better performance than most state-of-the-art methods.
%
\bibliographystyle{ACM-Reference-Format}
\bibliography{sample-base}


\end{document}